\documentclass[runningheads]{llncs}

\usepackage[T1]{fontenc}

\usepackage{graphics} 
\usepackage{epsfig} 
\usepackage{mathptmx} 
\usepackage{times} 
\usepackage{amsmath} 
\usepackage{amssymb}  
\usepackage{xcolor}
\usepackage{algorithm2e}
\usepackage{eso-pic}
\DeclareMathAlphabet{\mathcal}{OMS}{cmsy}{m}{n}
\DeclareSymbolFont{largesymbols}{OMX}{cmex}{m}{n}

\usepackage{comment}
\usepackage{graphicx}

\usepackage{balance}
\usepackage[utf8]{inputenc}
\usepackage[pdfauthor={C. Xu, E. Bonetto, A. Ahmad},
            pdftitle={DynaPix SLAM: A Pixel-Based Dynamic Visual SLAM Approach},
            pdfsubject={Dynamic Visual SLAM},
            pdfkeywords={Dynamic Visual SLAM, Moving Object Segmentation, Dynamic Environments}]{hyperref}
\usepackage{enumitem}
\usepackage{subcaption}
\usepackage{cuted}
\usepackage{multirow}
\hypersetup{
    colorlinks=true,
    linkcolor=blue,
    filecolor=magenta,      
    urlcolor=blue,
}

\usepackage{caption}

\usepackage{tikz}

\usepackage[textsize=tiny]{todonotes}

\newcommand{\uproman}[1]{\uppercase\expandafter{\romannumeral#1}}

\usepackage{soul}
\usepackage{array}
\usepackage{booktabs}
\newcolumntype{?}{!{\vrule width 2pt}}
\usepackage{cite}

\newif\ifreview
\reviewfalse

\ifreview
	\usepackage{lineno}

	\linenumbers
\fi

\begin{document}

\AddToShipoutPictureBG*{%
  \AtPageUpperLeft{%
    \setlength\unitlength{1in}%
    \hspace*{\dimexpr0.5\paperwidth\relax}
    \makebox(0,-1.5)[c]{\parbox{0.8\textwidth}{\centering This paper has been accepted for publication in the \\ 2024 DAGM German Conference on Pattern Recognition.\\
    Please cite as: Xu, C. and Bonetto, E. and Ahmad, A. (2024). DynaPix SLAM: A Pixel-Based Dynamic Visual SLAM Approach. \textit{DAGM German Conference on Pattern Recognition (GCPR) 2024}.}}%
}}

\def\SubNumber{70}
\def\GCPRTrack{Fast Review Track}

\title{DynaPix SLAM: A Pixel-Based Dynamic Visual SLAM Approach\thanks{The authors thank the International Max Planck Research School for Intelligent Systems (IMPRS-IS) for supporting Elia Bonetto.}}

\ifreview
\else
    \author{Chenghao Xu$^\dagger$\inst{1,3}\orcidID{0009-0007-2521-4493} 
    \and Elia Bonetto$^\dagger$\inst{2,3}\orcidID{0000-0003-0452-8761} 
    \and Aamir Ahmad\inst{2,3}\orcidID{0000-0002-0727-3031}}
    \authorrunning{C. Xu and E. Bonetto et al.}
    
    \institute{Swiss Federal Institute of Technology Lausanne (EPFL), Lausanne, Switzerland \email{chenghao.xu@epfl.ch}
    \and University of Stuttgart, 70569 Stuttgart, Germany \\
    \email{\{elia.bonetto, aamir.ahmad\}@ifr.uni-stuttgart.de}
    \and Max Planck Institute for Intelligent Systems, 72076 Tübingen, Germany}
\fi

\maketitle
\def\thefootnote{$\dagger$}\footnotetext{Chenghao Xu and Elia Bonetto contributed equally to this work as first authors.}

\begin{abstract}
Visual Simultaneous Localization and Mapping (V-SLAM) methods achieve remarkable performance in static environments, but face challenges in dynamic scenes where moving objects severely affect their core modules. To avoid this, dynamic V-SLAM approaches often leverage semantic information, geometric constraints, or optical flow. However, these methods are limited by imprecise estimations and their reliance on the accuracy of deep-learning models. Moreover, predefined thresholds for static/dynamic classification, the a-priori selection of dynamic object classes, and the inability to recognize unknown or unexpected moving objects, often degrade their performance. To address these limitations, we introduce DynaPix, a novel semantic-free V-SLAM system based on per-pixel motion probability estimation and an improved pose optimization process. The per-pixel motion probability is estimated using a static background differencing method on image data and optical flows computed on splatted frames. With DynaPix, we fully integrate these probabilities into map point selection and apply them through weighted bundle adjustment within the tracking and optimization modules of ORB-SLAM2. We thoroughly evaluate our method using the GRADE and TUM RGB-D datasets, showing significantly lower trajectory errors and longer tracking times in both static and dynamic sequences. The source code, datasets, and results are available at \url{https://dynapix.is.tue.mpg.de/}.

\end{abstract}    
\section{Introduction}
\setlength{\textfloatsep}{5pt}
\label{sec:intro}

\begin{figure}[!ht]
    \centering
    \includegraphics[width=\columnwidth]{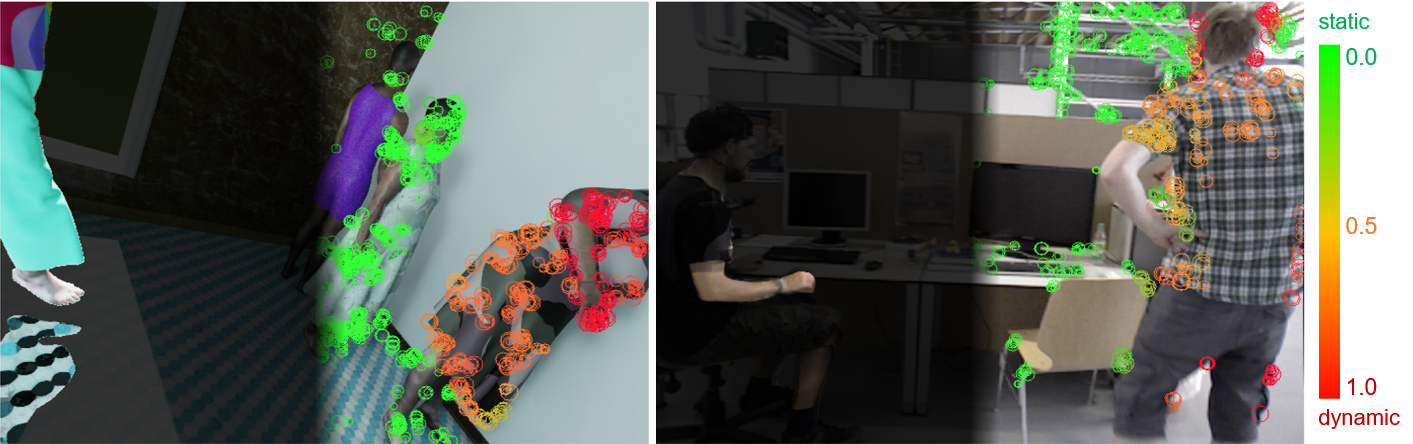}
    \caption{DynaPix's motion probabilities on GRADE (left) and TUM RGB-D (right) frames. On the left side of each image we colored the estimated \textit{moving} regions. On the right side, ORB features are colored based on the motion probabilities, from static (green) to dynamic (red).}
    \label{fig:preface}
\end{figure}

Visual SLAM algorithms have undergone significant development~\cite{7747236} and found wide-ranging applications in various scenarios, including service robots~\cite{service_SLAM}, autonomous vehicles~\cite{vehicle_SLAM}, and augmented reality devices~\cite{ar_SLAM}. However, most visual SLAM frameworks are developed under a \textit{static-world} assumption~\cite{vslam-survey}. The presence of dynamic objects violates such assumption and causes degradation in both \textbf{estimation accuracy and system robustness}~\cite{grade, robust-slam-survey, service_SLAM}, which limits their widespread deployment in real-world scenarios. Indeed, addressing this problem is necessary to develop robots that can safely act in dynamic environments. 

To resolve this issue, visual feature removal procedures have been introduced. Many are based on the detection or segmentation of dynamic object classes~\cite{dynaslam, DS-SLAM, DETECT-SLAM}. Several works instead use optical flow~\cite{flowfusion, flowangle2}, depth clustering~\cite{staticfusion, DCS}, or learning-based visual odometry methods~\cite{dytanvo, learning-seg-slam}. To increase robustness, other approaches combine geometric and semantic information~\cite{dynaslam, cfp-slam, SG-SLAM}, thus still heavily relying on semantic cues.
In general, all of those methods often suffer from failures related to prediction accuracy, noise, and/or imprecise estimations. Detection-based methods remove features belonging to the static environment within the detected bounding box. Segmentation is often approximate, especially around object borders or thin regions. Both detection- and segmentation-based methods have limited generalization capabilities, e.g. to different moving objects, as they rely on an a-priori choice of dynamic categories and the networks' performance. 
Furthermore, optical flow suffers from exploding magnitudes, especially in scenes with repeating patterns or featureless areas. Finally, learning-based methods often cannot generalize well to different cameras or scenarios~\cite{grade-ws,tartanair}. Moreover, most methods mask the \textit{entirety} of the objects without considering whether all or just part of them are moving — a common scenario for animals or humans. They also adopt simple binary moving/static masks instead of more informative blended probability distributions. This implies that \textit{all} features belonging to the identified moving region are rejected, independently of whether those are effectively moving or not. This can lead to frequent tracking losses and instability due to the loss of valuable information, especially when objects that might be moving occupy most of the image, or when there are many static features extracted from these objects that could be leveraged~\cite{grade,dytanvo}.



Therefore, we propose an innovative offline Visual SLAM system, that we call \textit{DynaPix}. Building on top of ORB-SLAM2~\cite{orb2}, DynaPix is based on a pixel-wise motion probability estimation technique for dynamic indoor environments.
The intuition is that the SLAM backend depends on a limited number of image features and keypoints associated with \textit{single} pixels. Thus, every extracted feature can contribute to the process through a weighting factor based on its motion state, regardless of whether it belongs to a \textit{supposedly} dynamic object or not. 
In the first stage, a novel static/dynamic differencing method on both neighboring frames and estimated optical flows is introduced. We use their outputs to compute per-pixel \textit{movable} and \textit{moving} probability estimations, respectively. These estimations are then combined to derive a per-pixel \textit{motion} probability that we use as a weighing factor in the SLAM backend, both within the map points selection and the weighted bundle adjustment (BA) procedures. To validate our method, we perform extensive evaluations and ablation studies on both the TUM RGB-D~\cite{tumrgbd} and the GRADE~\cite{grade} datasets, using the absolute trajectory error (ATE) metric and tracking rate for each experiment. 


To summarize, our main contributions are: i) the introduction of a novel per-pixel motion probability computation, without relying on semantic information, thresholds, or appearance-based detectors, ii) the integration of these probabilities in the tracking and optimization modules of a visual SLAM system, iii) rigorous experiments on public datasets showing that our approach achieves lower trajectory errors and higher tracking times in both synthetic and real-world sequence.


\begin{figure*}[!ht]
    \centering
    \includegraphics[width=\textwidth]{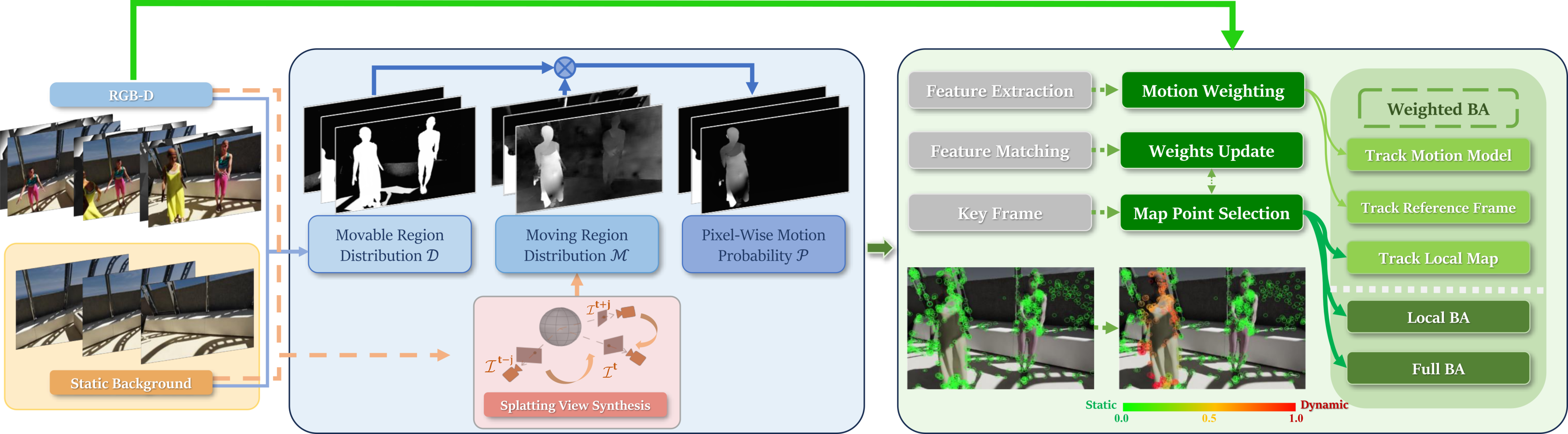}
    \caption{The DynaPix architecture consists of two main blocks, the \textit{motion} probability estimation (blue box), and the modified ORB-SLAM2 backend (green box). We compute \textit{movable} (Sec.~\ref{sec:movable}) and \textit{moving} regions (Sec.~\ref{sec:moving}) on the current frame. The per-pixel \textit{moving} probabilities (Sec.~\ref{sec:final_pm}) are then integrated into the SLAM backend (Sec.~\ref{sec:pose_opt}).}
    \label{fig:architecutre}
\end{figure*}
\section{Related Work}
\label{sec:related_work}
\subsubsection{Moving Object Segmentation} 
Typical approaches apply geometric constraints to eliminate dominant motion in images caused by camera movement, and then cluster regions that follow different rigid transformations~\cite{EMMS}. These methods often involve motion detection with depth clusters~\cite{3dmovobj, DCS}, multi-view epipolar geometry~\cite{epipolar}, or optical-flow-based methods~\cite{flowangle, flowangle2}. Learning-based flow estimation networks have also been investigated~\cite{PWC-Net, flownet}. 
However, these approaches often rely on predefined thresholds as detection criteria~\cite{rigidmask}, struggle to adapt to diverse environments~\cite{robust-slam-survey}, and suffer from inaccurate flow estimation. Deep Neural Networks (DNNs) can also be employed to detect or segment moving objects~\cite{dynaslam, DS-SLAM, DETECT-SLAM, dynamic-vins}.
However, those approaches require a set of \textit{fixed and predefined} classes. Therefore, they may mask completely stationary objects (e.g. parked cars), as well as the entirety of objects that are only partially moving~\cite{cfp-slam} (e.g. humans sitting). Moreover, they heavily rely on the chosen network's performance and can suffer from imprecise segmentation or incorrect object classifications. 

Methods combining geometric constraints, semantic information, and/or flow estimation 
apply a trade-off in identifying moving regions and are still typically based on learning-based frameworks. Therefore, they depend on their training data~\cite{flownet3d,rigidmask,cc}, aim to segment the scene into binary categories (moving/static), and mask entire object instances even if they are (partially) static. Moreover, these methods are often validated in outdoor scenarios, where most observations are static, thus limiting their applicability to dynamic indoor environments. In this work, we avoid considering only full objects and any predefined threshold or segmentation class to mark pixels as static/dynamic. Instead, we generate a moving probability value for each pixel by combining background differencing with optical flow information from splatted views.


\subsubsection{Dynamic SLAM} Most dynamic visual SLAM methods are \emph{de-facto} enhanced versions of classic SLAM frameworks~\cite{orb2, VINS} combined with motion segmentation techniques. Dynamic regions are either separately tracked~\cite{dymslam, DOT, dynaslam2} or discarded as outliers~\cite{dynaslam, DS-SLAM, DETECT-SLAM, dpslam, dynaremoval, DRG-SLAM} to reduce the negative effects on pose estimation. DynaSLAM~\cite{dynaslam} combines Mask R-CNN~\cite{maskrcnn} and multi-view geometry to process moving objects, DS-SLAM~\cite{DS-SLAM} applies a lightweight SegNet~\cite{SegNet} to obtain segmented masks, and Dynamic-VINS~\cite{dynamic-vins} uses YOLO-based detections. 
However, these methods suffer from wrongful detections, retain few features when there are many dynamic objects, and fail to identify moving objects outside the selected classes, thus leading to frequent tracking failures~\cite{grade,robust-slam-survey}.

Methods that integrate motion probability, like Detect-SLAM~\cite{DETECT-SLAM} which derives it from object detections, also fuse semantic attributes with geometric ones. DP-SLAM~\cite{dpslam} and Cheng et al.~\cite{dynaremoval} put forward dynamic region removal techniques within the Bayesian framework to enhance motion probability updates.
DE-SLAM~\cite{DESLAM} instead restores more static features for pose estimation of adjacent frames, even within selected dynamic categories. All of these, however, discard the dynamic features directly without retaining motion probabilities in the SLAM optimization process.
Close to our work,~\cite{cfp-slam, WF-SLAM, OVD-SLAM} use motion probability as weights in the BA procedure. However, their weights, used \textit{only} within the tracking module of ORB-SLAM2, are continuously discounted during the optimization process, making their effect negligible over subsequent iterations. Moreover, they often require additional optimization steps to attain acceptable results. In contrast, our method updates the motion probabilities following subsequent observations and propagates them throughout the entire SLAM framework.
In concurrent work, Liu et al.~\cite{Liu2023} proposed a method based on intensity change, sparse optical flow, and clustering to compute motion probabilities. However, their method heavily relies on thresholds due to the noise from those components, limits consideration to the current and the last frames, and uses intensity errors rather than our background reconstruction. Furthermore, similar to previous studies, these approaches often neglect to report the effective tracking time in their experiments. Indeed, the vast majority of evaluations focus only on trajectory errors (e.g. ATE and RPE). The total time a method tracks a trajectory is often overlooked, despite it being necessary for a thorough assessment of SLAM performance~\cite{robust-slam-survey, grade}.




\section{Approach}
\label{sec:approach}

Removing dynamic observations is a de-facto standard in dynamic visual SLAM, where many frameworks use a limited number of keypoints associated with specific image features. Then, while the at-unison movement of pixels can be captured as part of a supposedly dynamic object, e.g. with segmentation models, the information is still carried by the single feature located on a precise pixel of the image. Moreover, not always the entirety of an object is moving simultaneously, e.g. when a person walks or waves a hand. Thus, rather than grouping all pixels belonging to the same object equally, and inspired by optical flow-based methods, we approach the problem from a pixel-wise perspective. Furthermore, we note that directly removing the features from the system using a binary static/moving classification, without considering their current motion states, may produce a significant loss of valuable and usable information. Therefore, our goal is to estimate a movement probability associated with each pixel and dynamically integrate this notion into the SLAM backend. This allows us to selectively weigh singular feature points rather than binary classify whole groups of pixels equally.

As a result, our robust visual SLAM system for indoor dynamic environments consists of two novel modules: a pixel-wise motion probability estimator and an enhanced pose optimization process. The architecture of the proposed DynaPix SLAM system is outlined in Fig.~\ref{fig:architecutre}.
Initially, the system inputs are RGB-D sequences and corresponding static images. The static images represent the same scene in its version without any dynamic entity. Those can be generated either in simulation through synthetic generation or, for real-world sequences, by estimated background images, e.g. via video inpainting, view synthesis, or background filling techniques~\cite{e2fgvi,spinnerf,dynaslam}. Then, the motion estimator generates a \textbf{probabilistic representation} which identifies moving elements within the current image frame (Sec.~\ref{sec:motionprob}). This module enables the detection of \textbf{specific moving parts} within objects, as well as shadow and reflections, overcoming the limitations of traditional semantic or rigid motion segmentation methods.
Then, we incorporate these motion probabilities into the SLAM back-end through a \textbf{dynamic weighting} of the keypoints (Sec.~\ref{sec:pose_opt}). This refined optimization module ensures the preservation of more keypoints during the tracking phase, thereby enhancing the system's robustness with extended tracking duration.

\subsection{Pixel-wise Motion Probability}
\label{sec:motionprob}
To eliminate the influence of dynamic objects, we first estimate what is moving within the current scene. We introduce a two-staged method to identify i) potentially dynamic (\textit{movable}) and ii) currently \textit{moving} regions. The \textit{movable} regions are the area of the image where motion \textbf{can} occur. This serves as prior confidence to refine \textit{moving} region estimation into pixel-wise motion probabilities.

\subsubsection{Movable Region Estimation}
\label{sec:movable}
Although semantic cues are effective in identifying movable regions, they are incapable of capturing objects beyond predefined categories and imperceptible variations induced by dynamic objects, such as occlusions, reflections, and shadows. Thus, we use the static background image, $\mathcal{I}_{BG}$, in our system as prior information alongside the current image, $\mathcal{I}$. The difference between the current dynamic and static images in the RGB domain is captured by 
\begin{small}
    \begin{equation}
        \mathcal{I}_{d} = |\mathcal{I} - \mathcal{I}_{BG}|
    \end{equation}
\end{small}
\noindent Given the image difference $\mathcal{I}_{d}$, the per-pixel \textit{movable} probability $\mathcal{D}$ is then computed through:

\begin{small}
\begin{equation}
    \mathcal{D} = \lambda \cdot f_{1}(\mathcal{I}_{d}) + (1-\lambda) \cdot f_{2}(\mathcal{I}_{d}),
\label{eq:movable}
\end{equation}
\end{small}where $f_{1}$ and $f_{2}$ indicate respectively clipping and min-max normalizations to scale $\mathcal{I}_{d}$ into $[0, 1]$. The factor $\lambda$, given as 

\begin{small}
\begin{equation}
    \lambda = \frac{1}{2} + \frac{1}{\mbox{exp}(0.04 \cdot \mbox{max}(\mathcal{I}_{d}))+1}  \in [0.5, 1],
\end{equation}
\end{small}is used to weight these two terms. This is necessary because the direct application of $f_{2}$ (min-max normalization) on $\mathcal{I}_{d}$ could cause exploding scaling issues. Such problems typically occur when 
the current observation $\mathcal{I}$ closely resembles its static background $\mathcal{I}_{BG}$,  potentially due to the absence of dynamic objects coupled with minor color discrepancies.
This movable estimation process, illustrated in the supplementary material, can provide reliable information on all potential dynamic objects. Naturally, depending on how the background image is generated and what is considered to be movable or not, this step will have different effects as $\mathcal{I}_{BG}$ will vary accordingly.

\subsubsection{Moving Region Estimation}
\label{sec:moving}
Having identified the movable regions, we proceed to estimate the currently moving parts rather than segmenting entire objects.
The moving region can be decomposed into several sets of 3D points, each characterized by continuous displacements in Euclidean space. 
These displacements can be further projected onto the current 2D image frame for observations, resulting in pixel-wise motion across successive frames. 
To capture this, we first reproject adjacent frames into the current view to eliminate camera motion. We apply FlowFormer~\cite{flowformer} to calculate optical flow, thereby deriving pixel-wise displacements to identify moving regions. Ideally, when comparing the current frame with the reprojected frames, static pixels should remain at the same coordinates, while moving pixels display noticeable displacements. However, this approach can be compromised by incomplete elimination of camera motion or false pixel correspondences in texture-less areas. Therefore, as illustrated in the supplementary material, we adopt \textit{splatting-based view synthesis} for accurate projection, and \textit{static/dynamic flow differencing} for robust moving region estimation. 

\subsubsection{Splatting-based View Synthesis} 
\begin{figure}[!ht]
         \subcaptionbox{Frame $\mathcal{I}^{t}$}{\includegraphics[width=0.29\columnwidth]{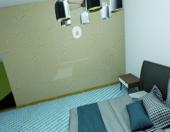}}\hfill
         \subcaptionbox{Homography TF}{\includegraphics[width=0.29\columnwidth]{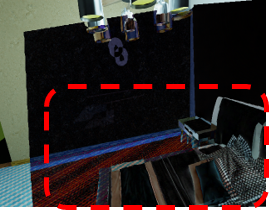}}\hfill
         \subcaptionbox{Softmax Splatting}{\includegraphics[width=0.29\columnwidth]{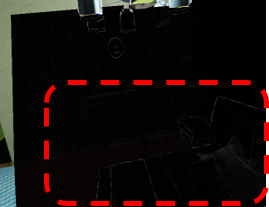}}\hfill
        \caption{Frame differences between a reprojected frame $\widetilde{\mathcal{I}}^{t+i}$ and Frame $\mathcal{I}^{t}$ using homography (b) and softmax splatting (c) transformations. The noise, evident in (b), is definitely reduced in (c).}
        \label{fig:reprojection}
\end{figure}

Homography transformation is commonly used to reproject images from other viewpoints to the current location~\cite{motionremoval}.  
However, assuming that all observed points lie on the same plane, regardless of their depth information, results in mismatches between the current frame $\mathcal{I}^{t}$ and the reprojection of the previous frame in the current viewpoint, $\widetilde{\mathcal{I}}^{t-1}$, as shown in Fig.~\ref{fig:reprojection}. To address this, we follow the idea of \textit{softmax splatting}~\cite{splatting} for more accurate view synthesis. Given the camera intrinsic matrix $K$ and the depth map of the previous frame, $Z^{t-1}$, we project the pixels of $\mathcal{I}^{t-1}$ into the 3D space to recover their 3D information. Then, we reproject these 3D points, identified by $\textbf{X}^{t-1}$, to the current view using an initially estimated transformation $\{R \in \mathfrak{so}^{3}, p\in \mathbb{R}^{3 \times 1}\}$:

\begin{small}
    \begin{equation}
\begin{split}
    & \textbf{X}^{t-1} = K^{-1}\textbf{x}^{t-1}Z^{t-1} \\
    & \widetilde{\textbf{x}}^{t-1} = K(R\textbf{X}^{t-1}+p)
\end{split}
\end{equation}
\end{small}

\noindent where $\textbf{x}$ denotes any 2D pixel coordinates $(u,v)$ in the image, and  $\widetilde{\textbf{x}}^{t-1}$ represents the projection of the 3D points when observed from the current pose, i.e. the $(u', v')$ pixels of $\widetilde{\mathcal{I}}^{t-1}$. With these correspondences, each pixel of $\widetilde{\mathcal{I}}^{t-1}$ is finally synthesized by participation from adjacent ones:

\begin{small}
\begin{equation}
    \widetilde{\mathcal{I}}^{t-1}(u', v') = \frac{\vec{\Sigma}(\mbox{exp}(z)\cdot \mathcal{I}^{t-1}(u, v))}{\vec{\Sigma}(\mbox{exp}(z))}
\end{equation}
\end{small}

\noindent where $\vec{\Sigma}(\cdot)$ denotes the sum of all contributing pixels from the original frame $\mathcal{I}^{t-1}$, $z$ is the pixel's depth, and $\mbox{exp}(z)$ serves as the weighting factor. More details can be found in \cite{splatting}. In Fig.~\ref{fig:reprojection}, we show the advantage of splatted views over homography transforms.
\subsubsection{Flow Differencing}
After obtaining the reprojected view, $\widetilde{\mathcal{I}}^{t-1}$, the estimation of moving regions is formulated as the flow estimation between ${\mathcal{I}}^{t}$ and $\widetilde{\mathcal{I}}^{t-1}$, indicated with $\mathcal{F}({\mathcal{I}}^{t},\widetilde{\mathcal{I}}^{t-1})$. The displacement in pixel-wise correspondences is quantified by the flow magnitude. However, this may not fully eliminate either the camera motion effects or the incorrect pixel correspondences that frequently occur in texture-less and ambiguous texture-rich areas. Therefore, we also perform flow estimation on static background images, using the procedure previously described, to further compensate for possible errors in these regions through:

\begin{small}
\begin{equation}
    \mathcal{F}^{*}(\mathcal{I}^{t},\widetilde{\mathcal{I}}^{t-1}) =  \mbox{min}(\mathcal{F}({\mathcal{I}}^{t},\widetilde{\mathcal{I}}^{t-1}), \mathcal{F}({\mathcal{I}}^{t},\widetilde{\mathcal{I}}^{t-1}) - \mathcal{F}({\mathcal{I}}_{BG}^{t},\widetilde{\mathcal{I}}_{BG}^{t-1}))\\
\end{equation}
\end{small}

\noindent A low-pass filter is then used to moderate the flow magnitude increase, due to subtraction in noisy estimations, and a min-max normalization operation is applied to project these values into $[0,1]$. Assuming the scene dynamics are consistent over a short interval, the motion attribute of each pixel should remain similar or vary minimally in neighboring frames. Without loss of generality, this can be extended to other frames closely spaced in time. Therefore, we finalize the \textit{moving region} estimation across multiple frames as:

\begin{small}
    \begin{equation}
        \mathcal{M} = \frac{1}{2n} \sum_{j \in J} \left( \mathcal{F}^{*}({\mathcal{I}}^{t},\widetilde{\mathcal{I}}^{t+j}) + \mathcal{F}^{*}(\mathcal{I}^{t},\widetilde{\mathcal{I}}^{t-j}) \right)
\label{eq:average}
\end{equation}
\end{small}

\noindent where $J$ is a set of time offsets with $n$ as its cardinality. In our implementation, we adopt $J = [2]$ (i.e., $n = 1$). As each index of $\mathcal{M}$ corresponds to the pixels of $\mathcal{I}^{t}$ and their values are between 0 and 1, $\mathcal{M}$ is also bounded between 0 and 1. A flow magnitude different than zero, in principle, indicates that a pixel has effectively moved between two frames. However, noise and incorrect matching may also result in such outcomes. Moreover, in general, the bigger the movement the higher will be the magnitude and the detrimental effect on the SLAM backend. Thus, we chose to treat this quantity as our per-pixel \textit{moving} probability factor of frame $\mathcal{I}^{t}$.

\subsubsection{Motion Probability}
\label{sec:final_pm}

The final \textit{motion} probability $\mathcal{P}$ for the current frame is defined by integrating the per-pixel moving probabilities $\mathcal{M}$ (Eq.~\ref{eq:average}), with the movable ones $\mathcal{D}$ (Eq.~\ref{eq:movable}), through pixel-wise multiplication:
\begin{small}
    \begin{equation}
   \mathcal{P} =  \mathcal{D} \odot \mathcal{M}
\label{eq:p_moving}
\end{equation}
\end{small}
\noindent This multi-step processing allows us to effectively reduce noise resulting from hallucinations (e.g. in the background inpainting) or wrong estimations (e.g. in the optical flow computation) by weighing the two factors, as shown by our ablation studies.


\subsection{Tracking and Pose Optimization}
\label{sec:pose_opt}
We incorporate estimated motion probabilities into the front-end \textit{tracking} and back-end \textit{pose optimization} process within the ORB-SLAM2 framework~\cite{orb2}.
Different from prior works~\cite{dynaslam, DS-SLAM, Ji}, our method seeks to use \textbf{all stationary objects} (or their \textbf{stationary parts}) to improve localization performance, while actively preventing their negative effects once they resume their motion.


\subsubsection{Local Tracking}
To prevent tracking failures, we seek to preserve more keypoints, even those associated with \textit{stationary movable} or \textit{slightly moving} objects, particularly in scenarios where movable objects dominate the frame. Another concern is that the temporarily stationary keypoints may start to move in future frames, disrupting the pose estimation process. Therefore, we first perform a coarse pose estimation (i.e., \textit{motion-only BA}) using \textbf{all keypoints and their respective weights} during the tracking thread. Then, thresholds $\gamma_{add}$ and $\gamma_{del}$ are set to ensure that only higher-confidence and static keypoints are selected as map points for fine pose optimization, (i.e., during \textit{full} or \textit{local BA}). Then, given a frame $\mathcal{I}^{t}$ and its estimated per-pixel motion probability $\mathcal{P}$, all keypoints $\mathcal{K}^{t}=\{k_1^{t}, \dots, k_n^{t}\}$ are assigned with motion probabilities $\mathcal{P}(k_i^t)$ based on their coordinates.
Upon identifying a current frame $\mathcal{I}^{t}$ as a keyframe, its keypoints $\mathcal{K}^{t}$ are considered \textit{candidate} map points. Among these, keypoints with $\mathcal{P}(k_{i}^{t}) \leq \gamma_{add}$ are promoted to map points $\mathcal{Y}=\{y_{j}|y_{j} \in \mathbb{R}^3, \mathcal{P}(y_{j}^{t}) = \mathcal{P}(k_{i}^{t})\}$. The motion probability for each map point is continually updated by matched keypoints from new frames. Any map points exhibiting motion, indicated by $\mathcal{P}(y_{j}^{t}) \geq \gamma_{del}$, will be eliminated from the map. In our implementation, we set $\gamma_{add} = 0.05$ and $\gamma_{del}=0.1$.

\subsubsection{Weighted Bundle Adjustment}
Inspired by~\cite{cfp-slam,WF-SLAM, OVD-SLAM}, the respective dynamic weights are integrated into the BA optimization process. The reprojection cost is formulated as:

\begin{small}
    \begin{equation}
\mathcal{C} = \mathop{\arg\min}\limits_{R,t}\frac{1}{2}\sum_{i=1}^{n} w_i || {x}_{i} - \pi(R {X}_{i} + p)||_{\rho}
\label{eq:ba}
\end{equation}
\end{small}

\noindent where $\Vert\cdot \Vert_{\rho}$ denotes the robust Huber cost function, $R$ and $p$ represent the camera's orientation and translation, respectively. $X_{i} \in \mathbb{R}^3$ indicates a 3D point in the scene, with $x_i$ as its matched 2D keypoint coordinates. The function $\pi{(\cdot)}$ is used for camera projection. Error term weights are determined by $w_i = 1 - \mathcal{P}(k_i^t)$, where a higher $w_i$ indicates a greater probability of the point being static. 
The coarse estimation applies this to solely refine camera poses with weights between $[0,1]$. For fine optimization, we simultaneously refine camera poses and map point locations using the same cost function described in Eq.~\ref{eq:ba}, with weights constrained to [1-$\gamma_{del}$, 1]. Both optimizations are solved using the Levenberg-Marquardt method within the \texttt{g2o} framework. Consequently, differently from previous approaches that just discarded features based on a binary classification, we retain more keypoints that can now provide a \textit{weighted} contribution to the pose optimization without being indiscriminately discarded. Notably, as seen above, if the keypoint or the feature \textit{becomes} dynamic it is completely removed from the system.

\section{Experiments \& Analysis}
\label{sec:exp}
We evaluate our system using sequences from both the TUM RGB-D~\cite{tumrgbd} and the recently released GRADE~\cite{grade} datasets. TUM RGB-D includes multiple indoor dynamic sequences, often adopted to benchmark visual SLAM approaches. Specifically, we use the four \textit{fr3/walking} sequences \textit{\{halfsphere, static, rpy, xyz\}}, with respective durations of 35.81, 24.83, 30.61, and 28.83 seconds. 
From GRADE, we adopt the 60-second-long \textit{D, DH, WO, WOH, S, SH, F, FH} experiments, as they represent long-term synthetic sequences for dynamic environments which have proven challenging for many dynamic SLAM methods~\cite{grade-ws}. \textit{S[H]} represents experiments recorded in static environments, \textit{D[H]} has additional moving people, \textit{F[H]} features randomly flying objects, and \textit{WO[H]} involves occlusions of the camera sensor. The \textit{H} indicates that the camera is kept horizontal throughout the entire sequence. All experiments are rendered both with and without the dynamic objects, i.e., the camera is positioned in the same location and the scene is rendered twice. 

\begin{figure}[!ht]
    \centering
    \includegraphics[width=\columnwidth]{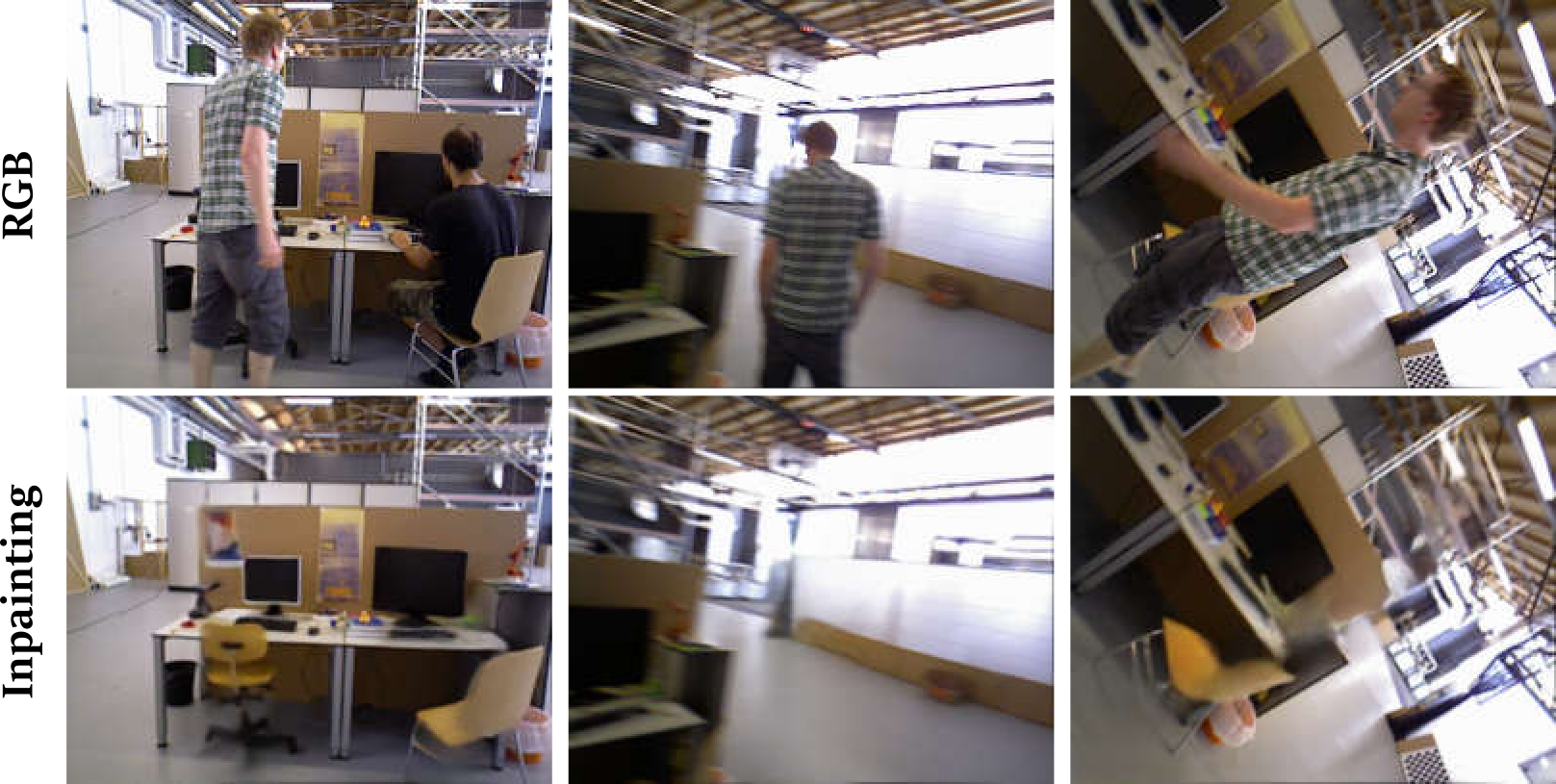}
    \caption{Examples of inpainted frames of the TUM RGB-D sequences. Those are then used in our \textit{movable} and \textit{moving} probability estimations.}
    \label{fig:inpainted}
\end{figure}

We use the E2FGVI~\cite{e2fgvi} video inpainting method to obtain static background images. To avoid failures on long-term sequences due to excessive GPU memory usage, we adapt our input frame strategy to a custom 50-frame sliding window approach and a 100-frame bootstrap with reference frames chosen based on camera poses rather than time intervals. Note that E2FGVI requires a mask input to select the inpainting region and we set it such that it covers the people in the frame. 
Examples of inpainted frames are illustrated in Fig.~\ref{fig:inpainted}. However, inpainting synthetic sequences is challenging due to the domain gap. Therefore, for those, we employ the available static sequences as static background images, i.e. without flying objects and people. Notably, the blurriness resulting from the hallucinations due to the inpainting process and varying lighting conditions in inpainted frames also introduce noise into our method and potentially reduce the number of available features. Due to the low tracking rate and high ATE of the original ORB-SLAM2 and DynaSLAM methods experienced in many GRADE sequences, we rely on the available camera poses for both the inpainting and splatting view synthesis. For consistency, we do so for both synthetic and real-world experiments. It is worth noting that neighboring \textit{frame-by-frame} pose variations can often be estimated or optimized through various visual (inertial) odometry or deep learning methods with high accuracy. We relax this constraint in our ablation studies for real-world sequences.

We benchmark DynaPix, including all components detailed in Sec.~\ref{sec:approach}, and DynaPix+, which extends DynaPix based on DynaSLAM~\cite{dynaslam}. DynaPix+ integrates semantic cues solely into \textit{map point selection}, allowing keypoints associated with dynamic classes to participate in tracking but not to be promoted as map points for subsequent fine optimization. We compare against their unmodified underlying methods, ORB-SLAM2 and DynaSLAM. Those are the main baseline methods since they utilize the same backend system, i.e. ORB-SLAM2. For completeness, we also evaluate DynamicVINS~\cite{dynamic-vins} (indicated with DynaVINS) and WF-SLAM~\cite{WF-SLAM} (for real-world sequences), although the use of different backends inherently makes those comparisons less informative. We perform ablation studies, presented in the supplementary material, on dynamic sequences to clarify the effect of each component with both DynaPix and DynaPix+. We use the RMSE of absolute trajectory error (ATE) and the tracking rate to measure accuracy and robustness. The TR represents the ratio of tracked time over the entire sequence duration, a metric often overlooked in many studies~\cite{grade, robust-slam-survey}. Although it is not a precise estimation of a future ATE forecast we also combine ATE and TR in a mixed metric, which we name \textbf{ATR}. This is obtained by computing the ratio $ATE/TR$. Clearly, the ATR will be negatively impacted by low TR and positively affected by low ATE, serving as an aggregating measure to compare the results. 
Our results on the GRADE dataset are reported in Tab.~\ref{tab:grade-all}, while the ones on the TUM RGB-D sequences are detailed in Tab.~\ref{tab:tum-full}. We repeat each experiment ten times using the default settings and report mean and standard deviation results. 
\begin{table*}[!ht]
        \newcolumntype{C}{>{\centering}p{3em}}
        \newcolumntype{d}{>{\centering}p{2em}}
    \centering
\caption{ATE RMSE [m] and Tracking Rate (TR) on both \textbf{static} and \textbf{dynamic} GRADE sequences. In bold the best ATR.}
    \resizebox{\linewidth}{!}{
    \begin{tabular}{lr|Cd|Cd|Cd|Cd|Cd?Cd|Cd|Cd|Cd|Cc}
    \multicolumn{2}{c|}{} & \multicolumn{10}{c?}{\textbf{GRADE -- Static}} & \multicolumn{10}{c}{\textbf{GRADE -- Dynamic}} \\ \cline{3-22}
    \multicolumn{2}{c|}{} & \multicolumn{2}{c|}{DynaPix} & \multicolumn{2}{c|}{ORB-SLAM2} & \multicolumn{2}{c|}{DynaPix+} & \multicolumn{2}{c|}{DynaSLAM}  & \multicolumn{2}{c?}{DynaVINS}  &     \multicolumn{2}{c|}{DynaPix} & \multicolumn{2}{c|}{ORB-SLAM2} & \multicolumn{2}{c|}{DynaPix+} & \multicolumn{2}{c|}{DynaSLAM}& \multicolumn{2}{c}{DynaVINS} \\ 
    \multicolumn{2}{c|}{} & ATE & TR & ATE & TR & ATE & TR & ATE & TR & ATE & TR & ATE & TR & ATE & TR & ATE & TR & ATE & TR & ATE & TR \\ \hline

    \multirow{2}{*}{FH} & mean & {0.006} & {1.00} & 0.010 & {1.00}  & 0.007 & {1.00} & 0.012 & {1.00} & 0.020 & 0.99 & {0.035} & {1.00} & 0.248 & {1.00} & 0.053 & {1.00} & 0.232 & 0.98 & 0.220 & 0.99 \\ 
                        & std & 0.000 & 0.00 & 0.002 & 0.00 & 0.000 & 0.00 & 0.005 & 0.00   & 0.002 & 0.00  & 0.016 & 0.00 & 0.103 & 0.00 & 0.029 & 0.00 & 0.041 & 0.02 & 0.058 & 0.00 \\ \hline
    \multirow{2}{*}{F} & mean & {0.230} & 0.86 & 0.330 & {0.90} & 0.323 & 0.88 & 0.529 & 0.82  & 1.176 & 0.83 & {0.291} & 0.20 & 0.359 & 0.35 & 0.579 & {0.64} & 0.864 & 0.42 & 1.532 & 0.84 \\ 
                        & std &0.426 & 0.00 & 0.507 & 0.01 & 0.505 & 0.01 & 0.526 & 0.04   & 0.404 & 0.26  & 0.108 & 0.01 & 0.156 & 0.03 & 0.465 & 0.23 & 0.217 & 0.07 & 0.504 & 0.23 \\ \hline
    \multirow{2}{*}{DH} & mean & {0.005} & {0.18} & {0.005} & {0.18} & 0.006 & {0.18} & 0.013 & 0.07  &  1.548 & 0.73 & 0.006 & {0.18} & {0.005} & {0.18} & {0.005} & {0.18} & 0.011 & 0.10 & 1.582 & 0.64 \\ 
                        & std &0.001 & 0.00 & 0.001 & 0.01 & 0.005 & 0.00 & 0.009 & 0.03   & 0.192 & 0.22  & 0.006 & 0.00 & 0.001 & 0.01 & 0.001 & 0.00 & 0.003 & 0.02 & 0.468 & 0.30 \\ \hline
    \multirow{2}{*}{D} & mean &0.023 & {0.98} & {0.018} & 0.97 & 0.019 & {0.98} & 0.024 & {0.98}  & 0.366 & 0.99 & {0.032} & {0.98} & 0.317 & 0.99 & 0.062 & 0.98 & 0.050 & 0.89 & 1.450 & 0.89 \\ 
                        & std & 0.006 & 0.01 & 0.002 & 0.03 & 0.005 & 0.00 & 0.008 & 0.02   & 0.330 & 0.00  & 0.010 & 0.02 & 0.043 & 0.01 & 0.043 & 0.03 & 0.009 & 0.05 & 0.441 & 0.10 \\ \hline
    \multirow{2}{*}{WOH} & mean & 0.012 & {0.54} & 0.013 & {0.54} & {0.008} & {0.54} & 0.015 & {0.54}  & 1.473 & 0.98 & {0.009} & {0.54} & 0.016 & {0.54} & 0.012 & {0.54} & 0.012 & {0.54} & 1.474 & 0.81 \\ 
                        & std & 0.008 & 0.00 & 0.010 & 0.00 & 0.004 & 0.00 & 0.017 & 0.00   & 0.264 & 0.01  & 0.001 & 0.00 & 0.008 & 0.00 & 0.002 & 0.00 & 0.002 & 0.00 & 0.548 & 0.12 \\ \hline
    \multirow{2}{*}{WO} & mean & 0.040 & 0.44 & {0.038} & {0.83} & 0.101 & 0.73 & 0.043 & 0.78  & 0.806 & 0.99 & {0.023} & {0.20} & 0.168 & {0.20} & 0.044 & {0.20} & 0.083 & 0.08 & 1.219 & 0.91 \\ 
                        & std & 0.023 & 0.35 & 0.021 & 0.32 & 0.202 & 0.34 & 0.022 & 0.28   & 0.442 & 0.00  & 0.002 & 0.00 & 0.022 & 0.00 & 0.006 & 0.00 & 0.010 & 0.00 & 0.291 & 0.09 \\ \hline
    \multirow{2}{*}{SH} & mean & 0.010 & {1.00} & 0.012 & {1.00}  & {0.009} & {1.00} & 0.010 & {1.00}  & 0.029 & 0.99 & --- & --- & --- & --- & --- & --- & --- & --- & --- & --- \\
                        & std & 0.001 & 0.00 & 0.002 & 0.00 & 0.001 & 0.00 & 0.001 & 0.00 & 0.005 & 0.00 & --- & --- & --- & --- & --- & --- & --- & --- & --- & --- \\ \hline
    \multirow{2}{*}{S} & mean & {0.010} & {1.00} & 0.011 & {1.00}  & 0.011 & {1.00} & {0.010} & {1.00} & 0.036 & 0.99 & --- & --- & --- & --- & --- & --- & --- & --- & --- & ---  \\
                        & std & 0.001 & 0.00 & 0.001 & 0.00 & 0.002 & 0.00 & 0.002 & 0.00  & 0.003 & 0.00 & --- & --- & --- & --- & --- & --- & --- & --- & --- & ---   \\ \hline
    \multicolumn{2}{c|}{Average} & {0.042} & 0.75 & 0.055 & {0.80} & 0.060 & 0.79 & 0.082 & 0.78 & 0.682 & 0.94 & {0.066} & 0.52 & 0.185 & 0.54 & 0.126 & {0.59} & 0.209 & 0.50 & 1.246 & 0.85 \\\hline
    \multicolumn{2}{c|}{ATR} & \multicolumn{2}{c|}{\textbf{0.056}} & \multicolumn{2}{c|}{0.069} & \multicolumn{2}{c|}{0.076} & \multicolumn{2}{c|}{0.105} & \multicolumn{2}{c?}{0.727} & \multicolumn{2}{c|}{\textbf{0.127}} & \multicolumn{2}{c|}{0.343} & \multicolumn{2}{c|}{0.214} & \multicolumn{2}{c|}{0.418} & \multicolumn{2}{c}{1.471} \\
    \end{tabular}
}
\label{tab:grade-all}
\end{table*}

\subsection{Static Sequences}
We use the static sequences from GRADE to demonstrate that DynaPix and DynaPix+ do not degrade the performance of their underlying SLAM frameworks in static scenes. Generally, as can be seen in Tab.~\ref{tab:grade-all}, both DynaPix and DynaPix+ perform well in such scenarios, often outperforming the base SLAM methods in most experiments. This is except for the \textit{WO} sequence, in which the TR of DynaPix is nearly half of the one with ORB-SLAM2. However, as indicated by the standard deviation, the significant variation in both ATE and TR for this sequence is due to a set of challenging featureless frames around the 26-second mark. This highlights the importance of using TR alongside ATE for comprehensive analysis of estimation results. Specifically, in the \textit{WO} experiment, the ATE remains similar in both methods, although the TR is significantly different. Focusing on the DynaPix+ and DynaSLAM results, our approach achieves better TR and ATE results overall, with $\sim26\%$ lower average ATE and a $\sim1\%$ increase of TR. These results also show how having a semantic segmentation system that blindly masks features out of the SLAM backend can adversely affect performance as a result of noise or incorrect detection, as DynaSLAM exhibits the lowest overall TR despite the static nature of these experiments. Finally, we notice how DynaVINS obtains a better tracking rate in the DH and WOH sequences but with at least eight times higher average ATE.

\subsection{Dynamic Sequences}
\begin{table*}[!ht]
    \centering
    \newcolumntype{C}{>{\centering}p{3em}}
    \newcolumntype{d}{>{\centering}p{2em}}
    
\caption{ATE RMSE [m] and Tracking Rate (TR) on the TUM RGB-D sequences. In bold the best ATR.}
    \resizebox{\linewidth}{!}{
    \begin{tabular}{lr|Cd|Cd?Cd|Cd?Cd|Cc}
    \multicolumn{2}{c|}{} & \multicolumn{2}{c|}{DynaPix} & \multicolumn{2}{c?}{ORB-SLAM2} & \multicolumn{2}{c|}{DynaPix+} & \multicolumn{2}{c?}{DynaSLAM} & \multicolumn{2}{c|}{WF-SLAM} & \multicolumn{2}{c}{DynaVINS} \\ \cline{3-14}
    \multicolumn{2}{c|}{} & ATE & TR & ATE & TR & ATE & TR & ATE & TR & ATE & TR & ATE & TR  \\ \hline
    \multirow{2}{*}{\/w\_half} & mean & 0.030 & {1.00} & 0.607 & 0.79 & {0.027} & {1.00} & 0.029 & {1.00} &  0.030 & 0.50 &  0.097 & 0.97\\ 
                               & std & 0.002 & 0.00 & 0.180 & 0.11 & 0.002 & 0.00 & 0.001 & 0.00 & 0.004 & 0.35 & 0.080 & 0.00   \\ \hline
    \multirow{2}{*}{\/w\_rpy} & mean & 0.043 & {1.00} & 0.744 & 0.99 & 0.038 & 0.99 & 0.040 & 0.86 & {0.031} & 0.59 & 0.130 & 0.97 \\  
                              & std & 0.007 & 0.00 & 0.115 & 0.01 & 0.007 & 0.02 & 0.008 & 0.03 & 0.007 & 0.07 & 0.027 & 0.01 \\ \hline
    \multirow{2}{*}{\/w\_static} & mean & 0.012 & {1.00} & 0.355 & {1.00} & 0.008 & {1.00} & {0.007} & 0.98 & {0.007} & 0.97 & 0.542 & 0.97 \\  
                              & std & 0.002 & 0.00 & 0.121 & 0.00 & 0.001 & 0.00 & 0.000 & 0.00 & 0.000 & 0.02  & 0.757 & 0.03 \\ \hline
    \multirow{2}{*}{\/w\_xyz} & mean & 0.018 & {1.00} & 0.732 & 0.84 & {0.015} & {1.00} & 0.016 & 0.92 & 0.017 & 0.99 & 0.055 & 0.97 \\  
                              & std & 0.002 & 0.00 & 0.102 & 0.11 & 0.000 & 0.00 & 0.001 & 0.00 & 0.001 & 0.02 & 0.014 & 0.00 \\ \hline
    \multicolumn{2}{c|}{Average} & 0.026 & {1.00} & 0.610 & 0.91 & 0.022 & {1.00} & 0.023 & 0.94 & {0.021} & 0.76 & 0.206 & 0.97\\\hline
    \multicolumn{2}{c|}{ATR} & \multicolumn{2}{c|}{0.026} & \multicolumn{2}{c?}{0.670} & \multicolumn{2}{c|}{\textbf{0.022}} & \multicolumn{2}{c?}{0.024} & \multicolumn{2}{c|}{0.028} & \multicolumn{2}{c}{0.212} \\ 
    \end{tabular}
}
\label{tab:tum-full}
\end{table*}

As shown in Tab.~\ref{tab:grade-all}, DynaPix and DynaPix+ consistently outperform their base methods by considerable amounts across both synthetic and real-world sequences. The TR of \textit{WO} with DynaPix+, for example, is 2.5 times better than DynaSLAM, with a lower ATE as well. For the same sequence, DynaPix shows a remarkable 86\% improvement in ATE over ORB-SLAM2, maintaining the \textit{same} TR. An exception is observed in the \textit{F} sequence with DynaPix, showing a 15\% shorter tracking time compared to ORB-SLAM2. This is attributed to the camera facing featureless walls, immediately followed by an area with multiple dynamic entities, causing the SLAM system to lose track in some trials. Nonetheless, average results from synthetic sequences demonstrate the superiority of DynaPix and DynaPix+ over unmodified methods, with up to 3 times lower ATE, and for DynaPix+ $\sim20\%$ higher TR. DynaVINS, probably thanks to the different backend system, achieves the best tracking rate (+25\%), at the expense of a ten times higher ATE when compared with DynaPix+. Overall, the best ATR, i.e. 0.127, is obtained by using DynaPix, with DynaPix+ being a close second with 0.214. DynaVINS, despite the higher TR, lags behind with an ATR of 1.471. Similarly, evaluations on TUM RGB-D (Tab.~\ref{tab:tum-full}) reveal DynaPix achieving 23 times ATE improvements over ORB-SLAM2, with 100\% TR, marking a 9.8\% improvement. DynaPix+ attains almost equal trajectory errors as DynaSLAM, but with a 6\% higher TR, primarily driven by 16\% and 8\% improvements in the \textit{rpy} and \textit{xyz} sequences, respectively. Meanwhile, WF-SLAM achieves the best overall ATE but is associated with the worst TR, approximately 24\%. DynaVINS, instead, cannot track the full trajectories, showing a TR of 97\%. Moreover, it also exhibits higher tracking rates in general. Therefore, its ATR is roughly ten times higher than the one obtained with DynaPix+. This further highlights the importance of not relying \textit{only} on the best ATE to validate the results of dynamic SLAM methods, but also investigating the tracking rate performance. Overall, these results indicate how retaining more features and keypoints, while dynamically weighting their contribution within the backend SLAM framework, can have a positive impact on both the ATE and the TR. An analysis of the importance of each component of the system and its impact on the performance is carried out and reported in our ablations studies in the supplementary material. There, we show how each step of the pipeline has a positive impact on the overall SLAM performance. 

\section{Conclusion}
We introduce DynaPix, an offline dynamic visual SLAM method that innovatively integrates pixel-wise motion probability estimation into a customized SLAM framework.
The dual-stage process blends \textit{movable} and \textit{moving} estimations from frame differencing and splatted optical-flow subtraction, respectively.
These motion probabilities are then employed in the tracking and backend optimization procedures of ORB-SLAM2 via a dynamic weighting mechanism. 
Through extensive testing on synthetic and real-world sequences, we underscore the importance of jointly analyzing trajectory errors and tracking rates in SLAM evaluation and showcase the superior performance of DynaPix and DynaPix+ over ORB-SLAM2, DynaSLAM, DynamicVINS, and WF-SLAM. 

Our findings highlight the advantages of distinguishing between static and moving parts of dynamic entities and weighting their influence based on motion status, thereby improving the robustness and accuracy of dynamic visual SLAM methods. 
Future works will focus on developing robust recovery strategies to address low tracking rates, achieving online and real-time performance by relaxing the reliance on future frames, optical flow, and inpainting techniques, and adapting our method to different SLAM backends.

\title{Supplementary Material for\\ DynaPix SLAM: A Pixel-Based Dynamic Visual SLAM Approach
}
 \author{Chenghao Xu$^\dagger$\inst{1,3}\orcidID{0009-0007-2521-4493} 
    \and Elia Bonetto$^\dagger$\inst{2,3}\orcidID{0000-0003-0452-8761} 
    \and Aamir Ahmad\inst{2,3}\orcidID{0000-0002-0727-3031}}
    \authorrunning{C. Xu and E. Bonetto et al.}
    
\begin{figure}[!ht]
    \centering
    \includegraphics[width=\columnwidth]{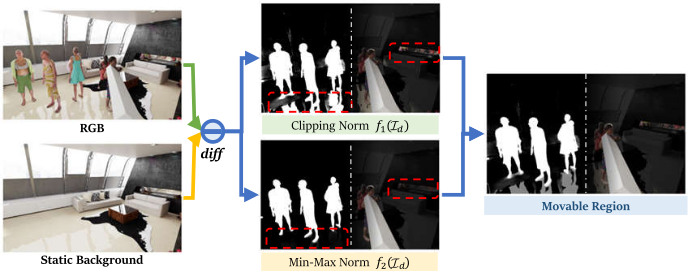}
    \caption{Example of \textit{movable} regions estimation.}
    \label{fig:movable}
\end{figure}

\begin{figure}[!ht]
    \centering \includegraphics[width=\columnwidth]{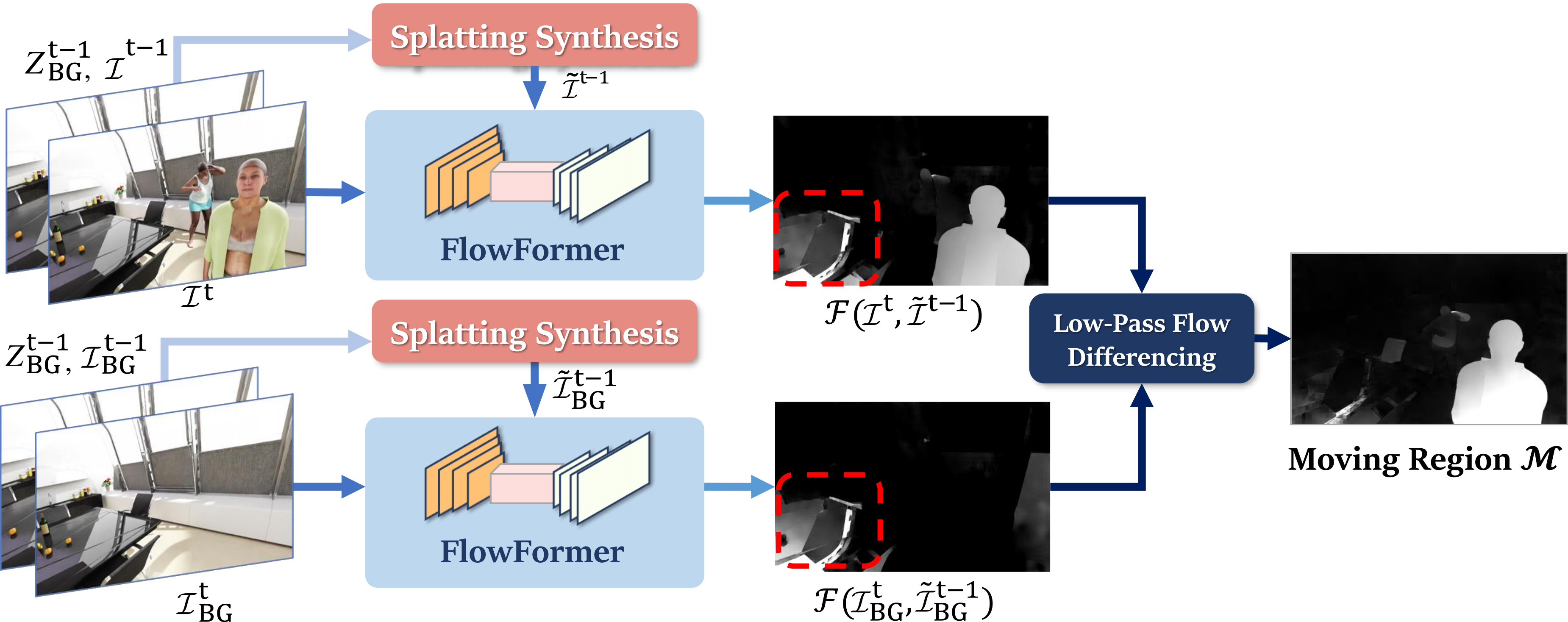}
    \caption{Example of \textit{moving} estimation between $\mathcal{I}^t$ and $\mathcal{I}^{t-1}$.}
    \label{fig:moving_module}
\end{figure}

    \institute{Swiss Federal Institute of Technology Lausanne (EPFL), Lausanne, Switzerland \email{chenghao.xu@epfl.ch}
    \and University of Stuttgart, 70569 Stuttgart, Germany \\
    \email{\{elia.bonetto, aamir.ahmad\}@ifr.uni-stuttgart.de}
    \and Max Planck Institute for Intelligent Systems, 72076 Tübingen, Germany}
       \titlerunning{DynaPix SLAM: A Pixel-Based Dynamic Visual SLAM Approach} \authorrunning{C. Xu and E. Bonetto et al.}
\maketitle
This is the supplementary material for GCPR 2024 paper ``DynaPix SLAM: A Pixel-Based Dynamic Visual SLAM Approach". We first present the schemes of our movable and moving estimations in Fig.~\ref{fig:movable} and Fig.~\ref{fig:moving_module}. We then present and discuss our ablation studies.

\section{Ablation Studies}
Our ablation studies for DynaPix and DynaPix+ on synthetic and real-world dynamic sequences are reported in Tab.~\ref{tab:grade-dyna-abl-full} and Tab.~\ref{tab:tum-abl-full}. The studies include i) disabling movable estimation, i.e., background differencing, ii) disabling moving estimation, i.e., optical flow component, iii) removing dynamic weighting, i.e., threshold-based keypoint selection without weighted BA, or iv) using estimated poses for both inpainting and splatting synthesis. The last aspect is applied only on TUM RGB-D sequences due to the lack of methods that achieve stable tracking for all GRADE sequences.

Disabling the movable estimation module affects both tracking rates and trajectory errors. On real-world sequences, DynaPix shows an average 35\% reduction in tracking rate, highlighting how relying solely on optical flow is insufficient. Conversely, DynaPix+ is more stable, largely thanks to the segmentation module from DynaSLAM. In synthetic sequences, particularly those with unknown flying objects, which pose challenges for pretrained detectors from \textit{FH} and \textit{F}, trajectory errors are the most affected by this ablation. 

Excluding the moving estimation module greatly impacts trajectory errors and experiment repeatability, with greater variances observed in both sets of experiments. The overall increase in trajectory errors indicates that only relying on removing all potentially moving elements from the scene is ineffective in all situations. Notably, in this scenario, DynaPix's TR for the \textit{FH} sequence drops to 0.03 (0.04 for DynaPix+), due to the numerous \textit{potentially} dynamic objects observed by the camera at the beginning of the experiment. This highlights again the importance of jointly analyzing ATE and TR, especially in complex scenarios.

We adopt standard BA optimization, i.e. without weights, and set a motion probability threshold of 0.05 for keypoints, i.e. features with higher values are excluded from the process. 
This significantly impacts trajectory errors in both experiment sets, suggesting a beneficial effect of integrating weights within the SLAM backend. Notably, DynaPix under this setting performs better on GRADE sequences, but with increased variances. Conversely, DynaPix+ demonstrates slightly lower trajectory errors and comparable tracking rates on TUM RGB-D, due to the segmentation module effectively removing the majority of keypoints associated with humans.

Ultimately, we use pose estimates from a previous run of DynaSLAM for splatting view synthesis and inpainting on TUM RGB-D sequences. For unavailable poses, such as during periods of non-tracking, we resort to the estimated poses from the nearest neighboring frames, where most errors can be reduced by \textit{flow differencing}. 

Overall, these experiments illustrate the positive effect of our introduced components on both tracking rate and trajectory error. This is also closely related to the specific characteristics of each testing sequence. Indeed, for scenes where humans are the only moving entities and positioned far from the camera, a simple segmentation method may suffice. However, the presence of additional moving objects, as seen in the \textit{F} and \textit{FH} sequences from GRADE, or failures in the detection network, as with rotated humans in \textit{rpy} sequence, necessitates a more generalized approach like DynaPix.
    \begin{table*}[!ht]
    \centering
\caption{ATE RMSE [m] and Tracking Rate (TR) of DynaPix and DynaPix+ ablation studies on the GRADE \textbf{dynamic} sequences.}
    \resizebox{\textwidth}{!}{
    \begin{tabular}{lr|cc|cc|cc?cc|cc|cc}
    \multicolumn{2}{c|}{} & \multicolumn{2}{c|}{\begin{tabular}{c} DynaPix \\ \textit{w/o movable est.}\end{tabular}} & \multicolumn{2}{c|}{\begin{tabular}{c} DynaPix \\ \textit{w/o moving est.}\end{tabular}} & \multicolumn{2}{c?}{\begin{tabular}{c} DynaPix \\ \textit{w/ threshold} \end{tabular}} & \multicolumn{2}{c|}{\begin{tabular}{c} DynaPix+ \\ \textit{w/o movable est.}\end{tabular}} & \multicolumn{2}{c|}{\begin{tabular}{c} DynaPix+ \\ \textit{w/o moving est.}\end{tabular}} & \multicolumn{2}{c}{\begin{tabular}{c} DynaPix+ \\ \textit{w/ threshold} \end{tabular}}\\
 \cline{3-14}
    \multicolumn{2}{c|}{} & ATE  & TR & ATE  & TR & ATE  & TR & ATE  & TR & ATE  & TR & ATE  & TR \\ \hline
    
    \multirow{2}{*}{FH} & mean & 0.285 & 1.00 & 0.042 & 0.03 & 0.138 & 1.00 & 0.103 & 1.00 & 0.040 & 0.04 & 0.135 & 1.00 \\ 
    & std & 0.127 & 0.00 & 0.007 & 0.00 & 0.136 &0.00 & 0.135 & 0.00 & 0.033 & 0.00 & 0.097 & 0.00 \\ \hline
    \multirow{2}{*}{F}  & mean & 0.454 & 0.32 & 0.295 & 0.23 & 0.149 &0.45 & 0.770 & 0.67 & 0.758 & 0.67 & 0.654 & 0.64 \\ 
    & std & 0.341 & 0.19 & 0.159 & 0.02 & 0.188 &0.24 & 0.459 & 0.18 & 0.460 & 0.18 & 0.491 & 0.22 \\ \hline
    \multirow{2}{*}{DH} & mean & 0.005 & 0.18 & 0.008 & 0.18 & 0.006 &0.18 & 0.004 & 0.18 & 0.005 & 0.18 & 0.007 & 0.18 \\ 
    & std & 0.000 & 0.01 & 0.004 & 0.01 & 0.002 &0.01 & 0.001 & 0.01 & 0.001 & 0.01 & 0.005 & 0.01 \\ \hline
    \multirow{2}{*}{D} & mean & 0.049 & 0.97 & 0.106 & 0.69 & 0.023 &0.99 & 0.038 & 0.98 & 0.079 & 0.86 & 0.051 & 0.99 \\ 
    & std & 0.055 & 0.06 & 0.066 & 0.14 & 0.004 &0.00 & 0.007 & 0.03 & 0.039 & 0.15 & 0.025 & 0.00 \\ \hline
    \multirow{2}{*}{WOH} & mean & 0.084 & 0.54 & 0.047 & 0.54 & 0.011 &0.54 & 0.061 & 0.54 & 0.021 & 0.54 & 0.012 & 0.54 \\ 
    & std & 0.038 & 0.00 & 0.017 & 0.00 & 0.002 &0.00 & 0.014 & 0.00 & 0.017 & 0.00 & 0.001 & 0.00 \\ \hline
    \multirow{2}{*}{WO} & mean & 0.040 & 0.20 & 0.065 & 0.20 & 0.029 &0.20 &  0.038 & 0.20 & 0.053 & 0.20 & 0.040 & 0.20 \\ 
    & std & 0.008 & 0.00 & 0.033 & 0.00 & 0.004 &0.00 & 0.005 & 0.00 & 0.005 & 0.00 & 0.005 & 0.00 \\ \hline
    \multicolumn{2}{c|}{Average} & 0.153 & 0.53 & 0.094 & 0.31 & 0.060 &0.56 & 0.169 & 0.59 & 0.159 & 0.41 & 0.150 & 0.59 \\ \hline
    \multicolumn{2}{c|}{ATR} & \multicolumn{2}{c|}{0.289} & \multicolumn{2}{c|}{0.303} & \multicolumn{2}{c?}{0.107} & \multicolumn{2}{c|}{0.286} & \multicolumn{2}{c|}{0.388} & \multicolumn{2}{c}{0.254}\\
    
    \end{tabular}
}
\label{tab:grade-dyna-abl-full}
\end{table*}

\begin{table*}[!ht]
    \centering
\caption{ATE RMSE [m] and Tracking Rate (TR) of DynaPix and DynaPix+ ablation studies on the TUM RGB-D Walking Sequences.}
    \resizebox{\textwidth}{!}{
    \begin{tabular}{lr|cc|cc|cc|cc?cc|cc|cc|cc}
    & & \multicolumn{2}{c|}{\begin{tabular}{c} DynaPix \\ \textit{w/o movable est.}\end{tabular}} & \multicolumn{2}{c|}{\begin{tabular}{c} DynaPix \\\textit{w/o moving est.}\end{tabular}} & \multicolumn{2}{c|}{\begin{tabular}{c} DynaPix \\ \textit{w/ threshold} \end{tabular}} & \multicolumn{2}{c?}{\begin{tabular}{c} DynaPix \\ \textit{w/ est. poses} \end{tabular}} & \multicolumn{2}{c|}{\begin{tabular}{c} DynaPix+ \\ \textit{w/o  movable est.}\end{tabular}} & \multicolumn{2}{c|}{\begin{tabular}{c} DynaPix+ \\ \textit{w/o moving est.}\end{tabular}} & \multicolumn{2}{c|}{\begin{tabular}{c} DynaPix+ \\ \textit{w/ threshold} \end{tabular}} & \multicolumn{2}{c}{\begin{tabular}{c} DynaPix+ \\ \textit{w/ est. poses} \end{tabular}} \\ \cline{3-18}
    \multicolumn{2}{c|}{} & ATE  & TR & ATE  & TR & ATE  & TR & ATE  & TR & ATE  & TR & ATE  & TR & ATE  & TR & ATE  & TR \\ \hline
    \multirow{2}{*}{\/w\_half} & mean & 0.053 & 0.31 & 0.214 & 1.00 & 0.096 & 1.00 & 0.091 & 1.00 & 0.026 & 1.00 & 0.031 & 1.00 & 0.026 & 1.00 & 0.028 & 1.00  \\    
                               & std & 0.028 & 0.04 & 0.076 & 0.29 & 0.089 & 0.36 & 0.090 & 0.36 & 0.001 & 0.00 & 0.009 & 0.00 & 0.001 & 0.00 & 0.002 & 0.00  \\ \hline 
    \multirow{2}{*}{\/w\_rpy} & mean & 0.046 & 0.64 & 0.206 & 0.99 & 0.050 & 0.97 & 0.074 & 0.99  & 0.034 & 0.90 & 0.041 & 1.00 & 0.029 & 0.96 & 0.038 & 1.00  \\     
                              & std & 0.013 & 0.09 & 0.168 & 0.16 & 0.162 & 0.18 & 0.155 & 0.19 & 0.004 & 0.01 & 0.006 & 0.00 & 0.005 & 0.03 & 0.004 & 0.01  \\ \hline 
    \multirow{2}{*}{\/w\_static} & mean & 0.011 & 1.00 & 0.153 & 1.00 & 0.010 & 1.00 & 0.013 & 1.00 & 0.007 & 1.00 & 0.013 & 1.00 & 0.007 & 1.00 & 0.008 & 1.00  \\     
                              & std & 0.001 & 0.00 & 0.058 & 0.00 & 0.067 & 0.00 & 0.082 & 0.00 & 0.000 & 0.00 & 0.011 & 0.00 & 0.001 & 0.00 & 0.001 & 0.00  \\ \hline 
    \multirow{2}{*}{\/w\_xyz} & mean & 0.019 & 0.90 & 0.052 & 1.00 & 0.167 & 1.00 & 0.026 & 1.00 & 0.015 & 1.00 & 0.015 & 1.00 & 0.015 & 1.00 & 0.015 & 1.00  \\     
                              & std & 0.001 & 0.01 & 0.022 & 0.04 & 0.022 & 0.05 & 0.022 & 0.05 & 0.001 & 0.00 & 0.000 & 0.00 & 0.000 & 0.00 & 0.001 & 0.00  \\ \hline 
    \multicolumn{2}{c|}{Average} & 0.032 & 0.65 & 0.156 & 1.00 & 0.081 & 0.99 & 0.051 & 1.00 & 0.021 & 0.98 & 0.025 & 1.00 & 0.019 & 0.99 & 0.022 & 1.00  \\   \hline
    \multicolumn{2}{c|}{ATR} & \multicolumn{2}{c|}{0.049} & \multicolumn{2}{c|}{0.156} & \multicolumn{2}{c|}{0.082} & \multicolumn{2}{c?}{0.051} & \multicolumn{2}{c|}{0.021} & \multicolumn{2}{c|}{0.025} & \multicolumn{2}{c|}{0.019} & \multicolumn{2}{c}{0.022}
    \end{tabular}
}
\label{tab:tum-abl-full}
\end{table*}

\bibliographystyle{splncs04}
\bibliography{070-main}

\end{document}